\documentclass[11pt,a4paper]{article}
\usepackage[hyperref]{naaclhlt2019}
\usepackage{times}
\usepackage{latexsym}
\usepackage{url}
\usepackage{amssymb}
\usepackage{amsmath}
\usepackage{graphicx}
\usepackage{caption}
\usepackage{subcaption}
\usepackage{booktabs}
\usepackage{tikz}
\usepackage{xcolor}
\usepackage{stfloats}
\usepackage[]{algpseudocode}
\usepackage{multirow}
\usepackage[normalem]{ulem}
\usepackage{enumitem}
\useunder{\uline}{\ul}{}

\aclfinalcopy 
\def\aclpaperid{1649} 

\newcommand{\svd}{{\texttt{SVD-aligned}}}
\newcommand{\adv}{{\texttt{ADV-aligned}}}

\tikzstyle{every picture}+=[remember picture]
\tikzstyle{none} = [shape=rectangle,inner sep=2pt,outer sep=1pt,text depth=0pt]
\tikzstyle{sentiment} = [shape=rectangle,inner sep=2pt,outer sep=1pt,text depth=7pt]
\tikzstyle{aspect} = [draw,shape=rectangle,inner sep=2pt,outer sep=1pt,text depth=0pt,fill=white!70!blue]
\tikzstyle{opinion} = [draw,dashed,line width=0.8pt,shape=rectangle,inner sep=2pt,outer sep=1pt,text depth=0pt,fill=white!50!red]
\tikzstyle{comment} = [shape=rectangle,inner sep=2pt,outer sep=1pt,text depth=0pt,scale=0.7]

\graphicspath{{images/}}

\title{Zero-Shot Cross-Lingual Opinion Target Extraction}

\author{Soufian Jebbara and Philipp Cimiano\\
  Semalytix GmbH, Bielefeld, Germany \\
    Semantic Computing Group, CITEC - Bielefeld University, Bielefeld, Germany \\
  \href{mailto:soufian.jebbara@semalytix.com}{soufian.jebbara@semalytix.com} \\
  \href{mailto:cimiano@cit-ec.uni-bielefeld.de}{cimiano@cit-ec.uni-bielefeld.de}
 \\}

\date{}

\begin{document}
\maketitle
\begin{abstract}
Aspect-based sentiment analysis involves the recognition of so called opinion target expressions (OTEs).
To automatically extract OTEs, supervised learning algorithms are usually employed which are trained on manually annotated corpora. The creation of these corpora is labor-intensive and sufficiently large datasets are therefore usually only available for a very narrow selection of languages and domains. 
In this work, we address the lack of available annotated data for specific languages by proposing a zero-shot cross-lingual approach for the extraction of opinion target expressions. We leverage multilingual word embeddings that share a common vector space across various languages and incorporate these into a convolutional neural network architecture for OTE extraction.
Our experiments with 5 languages give promising results: We can successfully train a model on annotated data of a source language and perform accurate prediction on a target language without ever using any annotated samples in that target language. Depending on the source and target language pairs, we reach performances in a zero-shot regime of up to 77\% of a model trained on target language data. Furthermore, we can increase this performance up to 87\% of a baseline model trained on target language data by performing cross-lingual learning from multiple source languages.
\end{abstract}

\section{Introduction}
In recent years, there has been an increasing interest in developing sentiment analysis models that predict sentiment at a more fine-grained level than at the level of a complete document.
A paradigm coined as Aspect-based Sentiment Analysis (ABSA) addresses this need by defining the sentiment expressed in a text relative to an opinion target (also called \emph{aspect}).
Consider the following example from a restaurant review:

\vspace{3mm}
\noindent
{%
\small
\textit{%
\tikz\node[none]{``};\tikz\node[aspect](a1){Moules}; \tikz\node[none]{were}; \tikz\node[opinion](s1){excellent};\tikz\node[none]{,}; \tikz\node[aspect](a2){lobster ravioli}; \tikz\node[none]{was}; \tikz\node[none]{VERY}; \tikz\node[opinion](s2){salty};\tikz\node[none]{!};\tikz\node[none]{''};%
}%
}
\begin{tikzpicture}[overlay]
  \path[->,black,semithick](s1) edge [out=160, in=20] (a1);
  \path[->,black,semithick](s2) edge [out=340, in=210] (a2);
\end{tikzpicture}
\vspace{3mm}

\noindent
In this example, there are two sentiment statements, one positive and one negative.
The positive one is indicated by the word ``\textit{excellent}'' and is expressed towards the opinion target ``\textit{Moules}''.
The second, negative sentiment, is indicated by the word ``\textit{salty}'' and is expressed towards the ``\textit{lobster ravioli}''.

A key task within this fine-grained sentiment analysis consists of identifying so called \emph{opinion target expressions} (OTE).
To automatically extract OTEs, supervised learning algorithms are usually employed which are trained on manually annotated corpora.
In this paper, we are concerned with how to transfer classifiers trained on one domain to another domain.
In particular, we focus on the transfer of models across languages to alleviate the need for multilingual training data.
We propose a model that is capable of accurate zero-shot cross-lingual OTE extraction, thus reducing the reliance on annotated data for every language.
Similar to \citet{upadhyay2018almost}, our model leverages multilingual word embeddings \cite{Smith2017offline,conneau2017word} that share a common vector space across various languages.
The shared space allows us to transfer a model trained on source language data to predict OTEs in a target language for which no (i.e.~zero-shot setting) or only small amounts of data are available, thus allowing to apply our model to under-resourced languages.

Our main contributions can be summarized as follows:
\begin{itemize}
    \item We present the first approach for zero-shot cross-lingual opinion target extraction and achieve up to 87\% of the performance of a monolingual baseline.
    \item We investigate the benefit of using multiple source languages for cross-lingual learning and show that we can improve by 6 to 8 points in F$_1$-Score compared to a model trained on a single source language.
    \item We investigate the benefit of augmenting the zero-shot approach with additional data points from the target language. We observe that we can save hundreds of annotated data points by employing a cross-lingual approach.
    \item We compare two methods for obtaining cross-lingual word embeddings on the task.
\end{itemize}

\section{Approach}
\label{sec:models}
A common approach for extracting opinion target expressions is to phrase the task as a sequence tagging problem using the well-known IOB scheme \cite{tksveenstra99eacl} to represent OTEs as a sequence of tags.
According to this scheme, each word in our text is marked with one of three tags, namely \textbf{I}, \textbf{O} or \textbf{B} that indicate if the word is at the \textbf{B}eginning\footnote{Note that the \textbf{B} token is only used to indicate the boundary of two consecutive phrases.}, \textbf{I}nside or \textbf{O}utside of a target expression. An example of such an encoding can be seen below:

\vspace{5mm}
{
    \small
    \begin{tabular}{cccccccc}
        \textit{The} & \textit{wine} & \textit{list} & \textit{is} & \textit{also} & \textit{really} & \textit{nice} & \textit{.} \\
        O & \textbf{I} & \textbf{I} & O & O & O & O & O 
    \end{tabular}
}
\vspace{5mm}

By rephrasing the task in this way, we can address it using established sequence tagging models. In this work, we use a multi-layer convolutional neural network (CNN) as our sequence tagging model.
The model receives a sequence of words as input features and predicts an output sequence of IOB tags.
In order to keep our model simple and our results clear, we restrict our input representation to a sequence of word embeddings.
While additional features such as Part-of-Speech (POS) tags are known to perform well in the domain of OTE extraction \cite{Toh2016,kumar2016iit,jebbara2016aspect}, they would require a separately trained model for POS-tag prediction which can not be assumed to be available for every language.
We refrain from using more complex architectures such as memory networks as our goal is mainly to investigate the possibility of performing zero-shot cross-lingual transfer learning for OTE prediction. Being the first approach proposing this, we leave the question of how to increase performance of the approach by using more complex architectures to future work.

In the following, we describe our monolingual CNN model for OTE extraction which we use as our baseline model.
Afterwards, we show how we adapt this model for a cross-lingual and even zero-shot regime.

\subsection{Monolingual Model}
\label{sec:mono-lingual-model}
Our monolingual baseline model consists of a word embedding layer, a stack of convolution layers, a standard feed-forward layer followed by a final output layer.
Formally, the word sequence $\mathbf{w}=(\mathbf{w}_1,\ldots,\mathbf{w}_n)$ is passed to the word embedding layer that maps each word $\mathbf{w}_i$ to its embedding vector $\mathbf{x}_i$ using an embedding matrix $\mathbf{W}$. 
The sequence of word embedding vectors $\mathbf{x}=(\mathbf{x}_1,\ldots,\mathbf{x}_n)$ is processed by a stack of $L$ convolutional layers\footnote{The input sequences are padded with zeros to allow the application of the convolution operations to the edge words.}, each with a kernel width of $l^{conv}$, $d^{conv}$ filter maps and RELU activation function $f$ \cite{Nair2010Rectified}.
The final output of these convolution layers is a sequence of abstract representations $\mathbf{h}^L=(\mathbf{h}^L_1,\ldots,\mathbf{h}^L_n)$ that incorporate the immediate context of each word by means of the learned convolution operations.
The hidden states $\mathbf{h}^L_i$ of the last convolution layer are processed by a regular feed-forward layer to further increase the model's capacity and the resulting sequence is passed to the output layer.

In a last step, each hidden state is projected to a probability distribution over all possible output tags $q_i=(q^B_i,q^I_i,q^O_i)$ using a standard feed-forward layer with weights $\mathbf{W}^{tag}$, bias $\mathbf{b}^{tag}$ and a softmax activation function.

\begin{figure}
  \centering
  \includegraphics[width=\columnwidth]{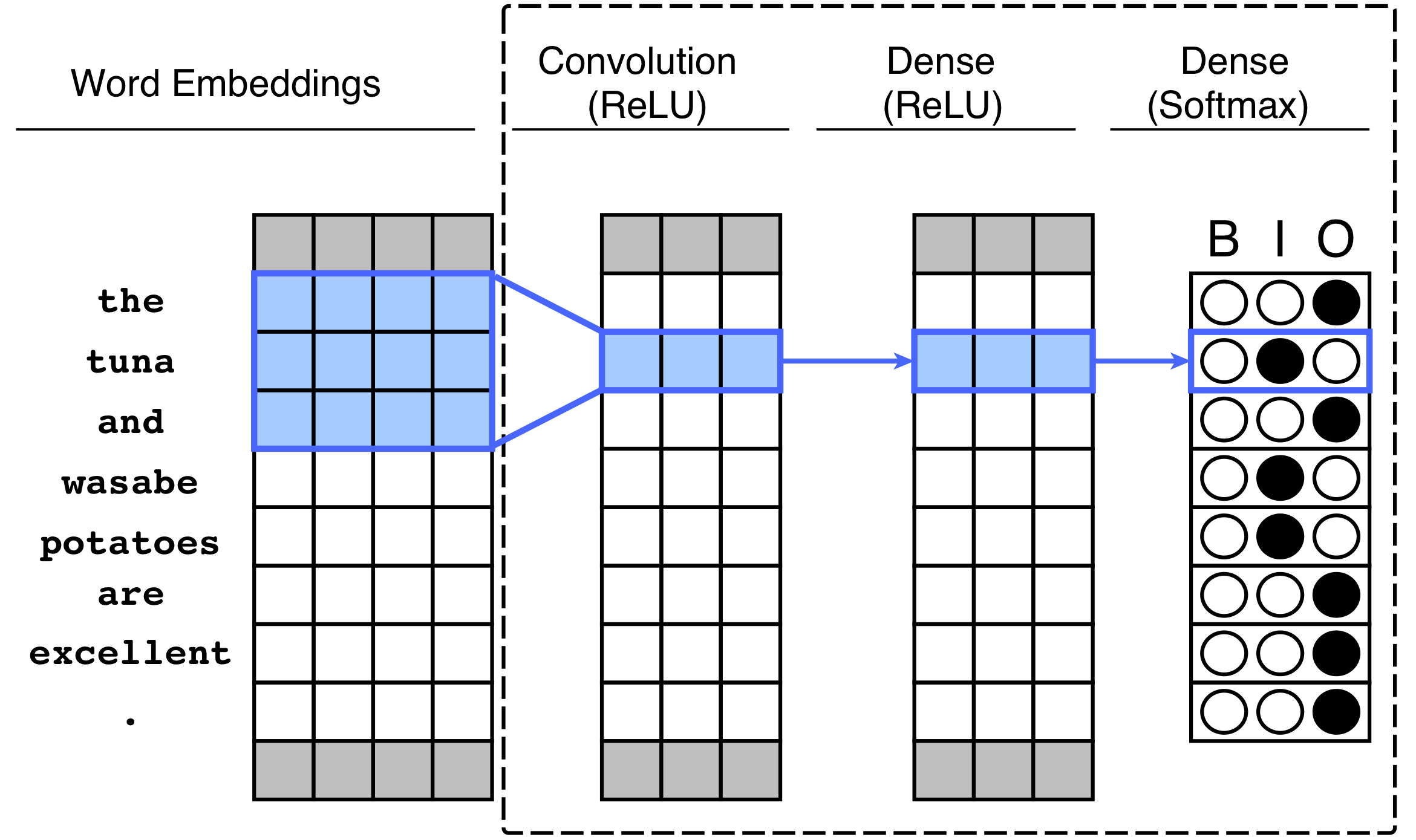}
  \caption{Model for sequence tagging using convolution operations. For simplicity, we only show a single convolution operation. The gray boxes depict padding vectors. The layers inside the dashed box are shared across multiple languages.}
  \label{fig:model:mono-lingual}
\end{figure}

Since the prediction of each tag can be interpreted as a classification, the network is trained to minimize the categorical cross-entropy between expected tag distribution $p_i$ and predicted tag distribution $q_i$ of each word $i$:
$$H(p_i,q_i) = -  \sum_{t \in \mathcal{T}} p_i^t \log(q_i^t),$$
where $\mathcal{T}=\{I,O, B\}$ is the set of IOB tags, $p_i^t \in \{0,1\}$ is the expected probability of tag $t$ and $q_i^t\in [0,1]$ the predicted probability.
Figure~\ref{fig:model:mono-lingual} depicts the sequence labeling architecture.

\subsection{Cross-Lingual Model}
\label{sec:cross-lingual-model}

Our cross-lingual model works purely with cross-lingual embeddings that have been trained on monolingual datasets and in a second step have been aligned across languages. 
In fact, the embeddings are pre-computed in an offline fashion and are not adapted while training the convolutional network on data from a specific language. 
As the inputs to the convolutional network are only the cross-lingual embeddings, the network can be applied to any language for which the embeddings have been aligned.

Since the word embeddings for source and target language share a common vector space, the shared parts of the target language model are able to process data samples from the completely unseen target language and perform accurate prediction i.e.~enabling zero-shot cross-lingual extraction of opinion target expressions.

We rely on two approaches to compute embeddings that are aligned across languages.
Both methods rely on fastText \cite{bojanowski2017enriching} to compute monolingual embeddings trained on Wikipedia articles.
The first method is the one proposed by \citet{Smith2017offline}, which computes a singular value decomposition (SVD) on a dictionary of translated word pairs to obtain an optimal, orthogonal projection matrix from one space into the other.
We refer to this method as \svd{}.
We use these embeddings\footnote{Obtained from: \url{https://github.com/Babylonpartners/fastText_multilingual}} in our experiments in Sections~\ref{sec:zero},~\ref{sec:eval:target} and~\ref{sec:sota}.

The second method proposed by \citet{conneau2017word} performs the alignment of embeddings across languages in an unsupervised fashion, without requiring translation pairs.

The approach uses adversarial training to initialize the cross-lingual mapping and a synthetically generated bilingual dictionary to fine-tune it with the Procrustes algorithm \cite{Schönemann1966}.
We refer to the multilingual embeddings\footnote{Obtained from: \url{https://github.com/facebookresearch/MUSE}} from \citet{conneau2017word} as \adv{}.
These are used in Section~\ref{sec:eval:diff}.

\section{Evaluation}
\label{sec:experiments}
In this section, we investigate the proposed zero-shot cross-lingual approach and evaluate it on the widely used dataset of Task 5 of the SemEval 2016 workshop.
With our evaluation, we answer the following research questions:
\begin{enumerate}[label={RQ\arabic*:}, align=left, leftmargin=*]
    \item To what degree is the model capable of performing OTE extraction for unseen languages?
    \item Is there a benefit in training on more than one source language?
    \item What improvements can be expected when a small amount of samples for the target language are available?
    \item How big is the impact of the used alignment method on the OTE extraction performance?
\end{enumerate}
Before we answer these questions, we give a brief overview over the used datasets and resources.

\subsection{Datasets}
\label{sec:dataset}
As part of Task 5 of the SemEval 2016 workshop \cite{Pontiki2016}, a collection of datasets for aspect-based sentiment analysis on various languages and domains was published.
Due to its relatively large number of samples and high coverage of languages and domains, the datasets are commonly used to evaluate ABSA approaches.
To answer our research questions, we make use of a selection of the available datasets.
We evaluate our cross-lingual approach on the available datasets for the restaurant domain for the 5 languages Dutch (\texttt{nl}), English (\texttt{en}), Russian (\texttt{ru}), Spanish (\texttt{es}) and Turkish (\texttt{tr})\footnote{We tried to include the dataset of French reviews in our evaluation but the provided download script no longer works.}.
Table~\ref{tab:datasets} gives a brief overview of the used datasets.
\begin{table}[]
    \centering
\begin{tabular}{lrrr}
\toprule
         Dataset &  \#Sent. &  \#Tokens &  \#Targets \\
\midrule
 \texttt{en} (train) &        2000 &    29278 &      1880 \\
 \texttt{en} (test) &         676 &    10080 &       650 \\
 \texttt{es} (train) &        2070 &    36164 &      1937 \\
 \texttt{es} (test) &         881 &    13290 &       731 \\
   \texttt{nl} (train) &        1722 &    24981 &      1283 \\
   \texttt{nl} (test) &         575 &     7690 &       394 \\
 \texttt{ru} (train) &        3655 &    53734 &      3159 \\
 \texttt{ru} (test) &        1209 &    17856 &       972 \\
 \texttt{tr} (train) &        1232 &    12702 &      1385 \\
 \texttt{tr} (test) &         144 &     1360 &       159 \\
\bottomrule
\end{tabular}
    \caption{Statistics of the SemEval 2016 ABSA dataset for the restaurant domain.}
    \label{tab:datasets}
\end{table}

\subsection{Experimental Settings}
In all our experiments, we report F$_1$-scores for the extracted opinion target expressions computed on exact matches of the character spans as in the original SemEval task \cite{Pontiki2016}.

As described in Section~\ref{sec:cross-lingual-model}, our model relies on pretrained multilingual embeddings.
For both \svd{} and \adv{}, we use the embeddings as provided by the original authors.
However, we restrict our vocabulary to the most frequent 50,000 words per language\footnote{As appearing in the respective embedding files.} to reduce memory consumption.

For all experiments, we fix our model architecture to 5 convolution layers with each having a kernel size of 3, a dimensionality of 300 units and a ReLU activation function \cite{Nair2010Rectified}.
The penultimate feed-forward layer has 300 dimensions and a ReLU activation, as well.
We apply dropout \cite{Hinton2014} on the word embedding layer with a rate of 0.3 and between all other layers with 0.5.
The word embeddings and the penultimate layer are L1-regularized \cite{Ng2004Feature}.

The network's parameters are optimized using the stochastic optimization technique \emph{Adam} \cite{Kingma2014}.
We optimize the number of training epochs for each model using early stopping \cite{CaruanaLG00} but do not tune other hyperparameters of our models.
We always pick 20\% of our available training data for the validation process.
For the zero-shot scenario, this entails that we optimize the number of epochs on the source language and not on the target language to simulate true zero-shot learning.

\subsection{Zero-Shot Transfer Learning}
\label{sec:zero}
In this section, we present our evaluation for zero-shot learning.
We first examine a setting with a single source language.
Then, we evaluate the effect of cross-lingual learning from multiple source languages.

\begin{figure}
  \centering
  \includegraphics[width=\columnwidth]{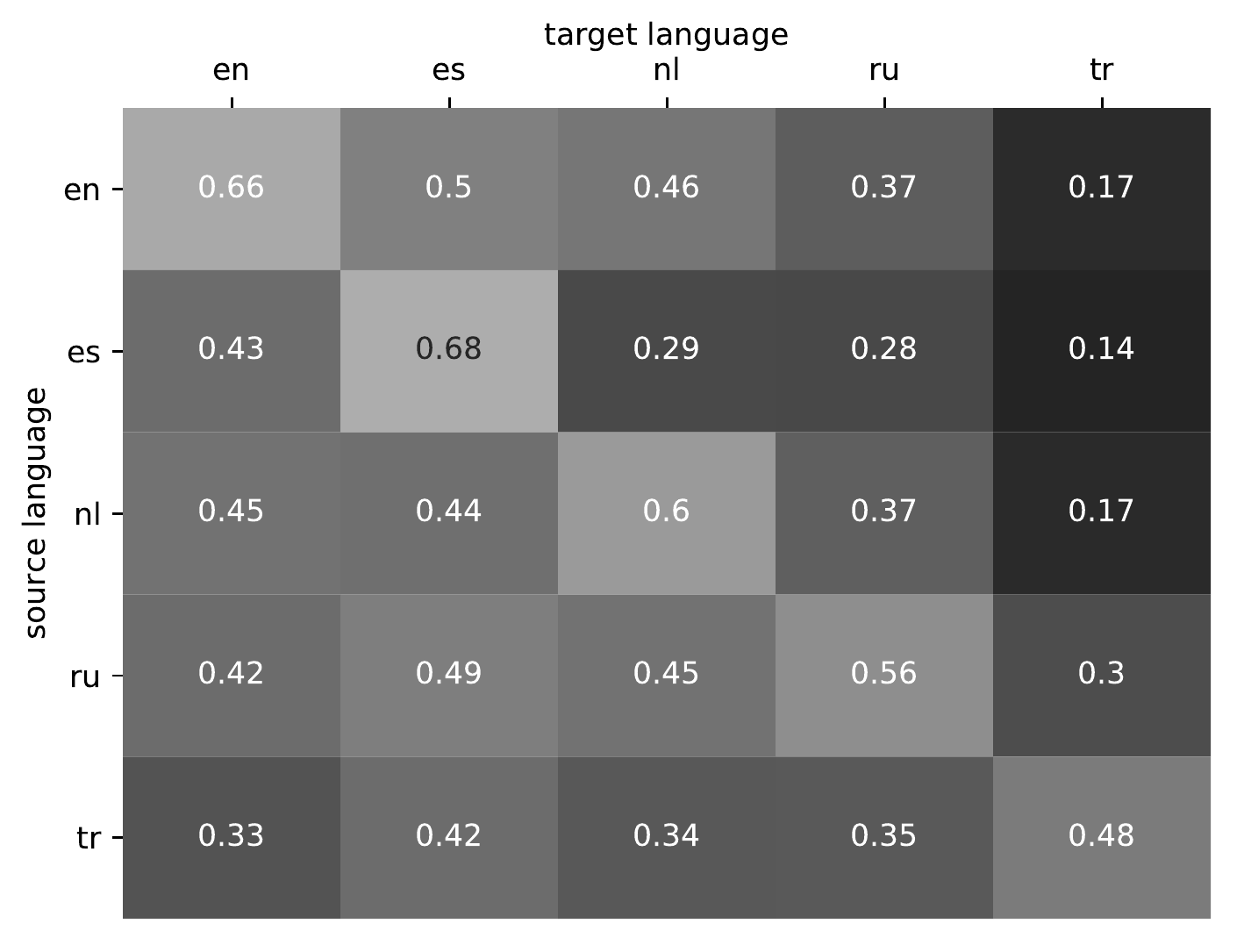}
  \caption{Zero-shot F$_1$-scores for cross-lingual learning from a single source to a target language.}
  \label{fig:single-zero}
\end{figure}

\paragraph{Single Source Language}
\label{sec:eval:zero-single}
This part of our evaluation addresses our first research question:
\begin{enumerate}[label={RQ1:}, align=left, leftmargin=*]
\item To what degree is the model capable of performing OTE extraction for unseen languages?
\end{enumerate}
To answer this question, we perform a set of experiments in the zero-shot setting.
We train a model on the training portion of a source language and evaluate the model performance on all possible target languages.
Figure~\ref{fig:single-zero} shows the obtained scores.
The reported results are averaged over 10 runs with different random seeds.
The main diagonal represents results of models both trained and tested on target language data.
We considered these our monolingual baselines.

In general, the proposed approach achieves relatively high scores for some language pairs,
although with large performance differences depending on the exact source and target language pairs.
Looking at the absolute scores, the best performing cross-lingual language pair is \texttt{en$\rightarrow$es} with an F$_1$-score of 0.5.
This is followed by \texttt{en$\rightarrow$nl} at 0.46.
The lowest is \texttt{es$\rightarrow$tr} with an F$_1$-score of 0.14.
When considering the results relative to their respective monolingual baselines, the highest relative performance is achieved by \texttt{en$\rightarrow$nl} at 77\% of a \texttt{nl$\rightarrow$nl} model, followed by \texttt{en$\rightarrow$es} and \texttt{ru$\rightarrow$nl}, which both reach an F-Measure of about 74\%.
The weakest performing language pair is still \texttt{es$\rightarrow$tr} at 29\% relative performance.
In general, the Turkish language seems to benefit the least from the cross-lingual transfer learning, while Russian is on average the best source language in terms of relative performance achievement for the target languages.

Overall, the presented results show that it is in fact possible for most considered languages to train a model for OTE extraction without ever using any annotated data in that target language.

\paragraph{Multiple Source Languages}
\label{sec:eval:zero-multiple}
In the next experiment, we want to address our second research question:
\begin{enumerate}[label={RQ2:}, align=left, leftmargin=*]
\item Is there a benefit in training on more than one source language?
\end{enumerate}
As we explained in Section~\ref{sec:cross-lingual-model}, our approach allows us to train and test on any number of source and target languages, provided that we have aligned word embeddings for each considered language.

In order to answer our second research question, we train a model on the available training data for all but one language and perform prediction on the test data for the left-out language.
The results for these experiments are summarized in Table~\ref{tab:multi-zero}.
We can see that all languages with the exception of Turkish seem to profit from a cross-lingual transfer setting with multiple source languages.
The absolute improvements are in the range of 6 to 8 points in F$_1$-Score while the performance on Turkish samples drops by 3 points.

\begin{table}[]
\small
\begin{tabular}{rrrrrr}
\toprule
                target &    en &    es &    nl &    ru &    tr \\
\midrule
       best$\rightarrow$target &  0.45 &  0.50 &  0.46 &  0.37 &  0.30 \\
 all others$\rightarrow$target &  0.52 &  0.58 &  0.53 &  0.43 &  0.27 \\
     target$\rightarrow$target &  0.66 &  0.68 &  0.60 &  0.56 &  0.48 \\
\bottomrule
\end{tabular}
  \caption{Zero-shot results for cross-lingual learning from multiple source languages to a target language. The row \texttt{best$\rightarrow$target} represents the best performing cross-lingual model from Figure~\ref{fig:single-zero} for each target language. \texttt{all others$\rightarrow$target} are the results for training on all languages except for the target language. \texttt{target$\rightarrow$target} shows the monolingual scores that act as a baseline.}
  \label{tab:multi-zero}
\end{table}
We can summarize that we can obtain substantial improvements for most languages when training on a combination of multiple source languages.
In fact, for \texttt{en}, \texttt{es}, \texttt{nl} and \texttt{ru}, the results of our cross-lingual models trained on all other languages reach between 78\% to 87\% relative performance of a model trained with target language data.

\subsection{Cross-Lingual Transfer Learning with Additional Target Language Data}
\label{sec:eval:target}
\begin{figure}
  \centering
  \includegraphics[width=\columnwidth]{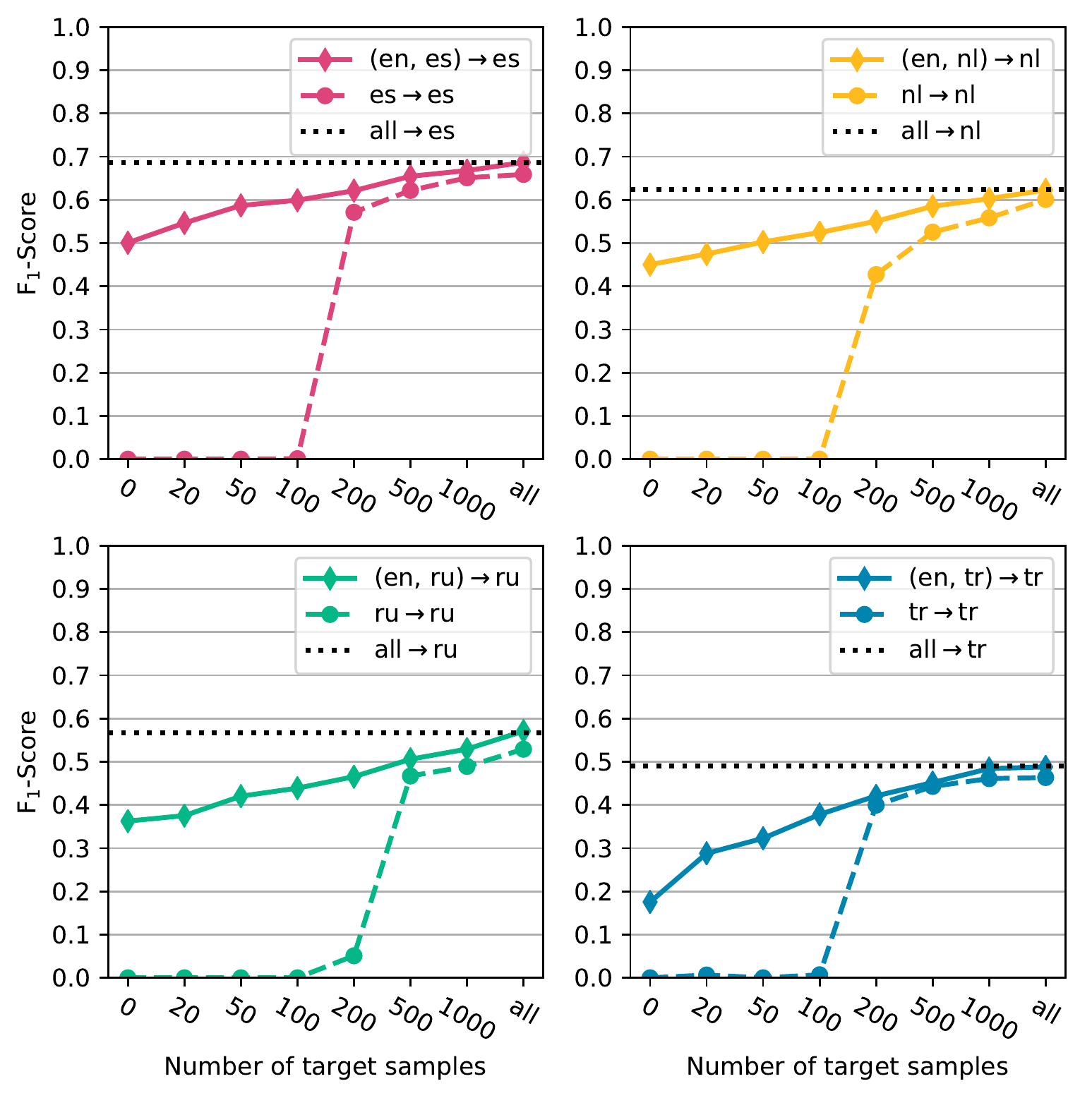}
  \caption{Cross-lingual results for increasing numbers of training samples from the target language.}
  \label{fig:target-data}
\end{figure}

While our goal is to reduce the effort of annotating huge amounts of data in a target language to which the model is to be transferred, it might still be reasonable to provide a few annotated samples for a target language.
Our next research question addresses this issue:
\begin{enumerate}[label={RQ3:}, align=left, leftmargin=*]
    \item What improvements can be expected when a small amount of samples for the target language are available?
\end{enumerate}
We answer this question by training our models jointly on a source language dataset as well as a small amount of target language samples and compare this to a baseline model that only uses target language samples.
By gradually increasing the available target samples, we can directly observe their benefit on the test performance.
Figure~\ref{fig:target-data} shows a visualization for the source language \texttt{en} and the target languages \texttt{es}, \texttt{nl}, \texttt{ru}, and \texttt{tr}.

We can immediately see that a monolingual model requires at least 100 target samples to produce meaningful results as opposed to a cross-lingual model that performs well with source language samples alone.
Training on increasing amounts of target samples improves the model performances monotonically for each target language and the model leveraging the bilingual data consistently outperforms the monolingual baseline model.
The benefits of the source language data are especially pronounced when very few target samples are available, i.e.~less than 200.
As an example, a model trained on bilingual data using all available English samples and 200 Dutch samples is competitive to a monolingual model trained on 1000 Dutch samples (0.55 vs. 0.56).

As one would expect, the results in Table~\ref{tab:multi-zero} and Figure~\ref{fig:target-data} suggest that training the model on more data samples leads to a better performance.
Since our model can leverage the data from all languages simultaneously, we can exhaust our resources and train an instance of our model that has access to all training data samples from all languages, including the target training data.
This is reflected by the dashed line in Figure~\ref{fig:target-data}.
We see, however, that the model cannot leverage the other source languages beyond what it achieves with the combination of the full target and English language data alone.

\subsection{Comparison of Alignment Methods}
\label{sec:eval:diff}
The previous experiments show that we can achieve good performance in a cross-lingual setting for OTE extraction using the multilingual word embeddings proposed by \citet{Smith2017offline}.
Now we address our final research question:
\begin{enumerate}[label={RQ4:}, align=left, leftmargin=*]
    \item How big is the impact of the used alignment method on the OTE extraction performance?
\end{enumerate}
With our final research question, we compare our previous results to an alternative method of aligning word embeddings in multiple languages.
We repeat our experiments in Section~\ref{sec:eval:zero-single} using the embeddings of \citet{conneau2017word} which we refer to as \adv{}.

To enable a direct comparison to the zero-shot results in Section~\ref{sec:eval:zero-single}, we report absolute differences in F$_1$-Score to the scores obtained with \svd{} for all source and target language combinations.

\begin{figure}
  \centering
  \includegraphics[width=\columnwidth]{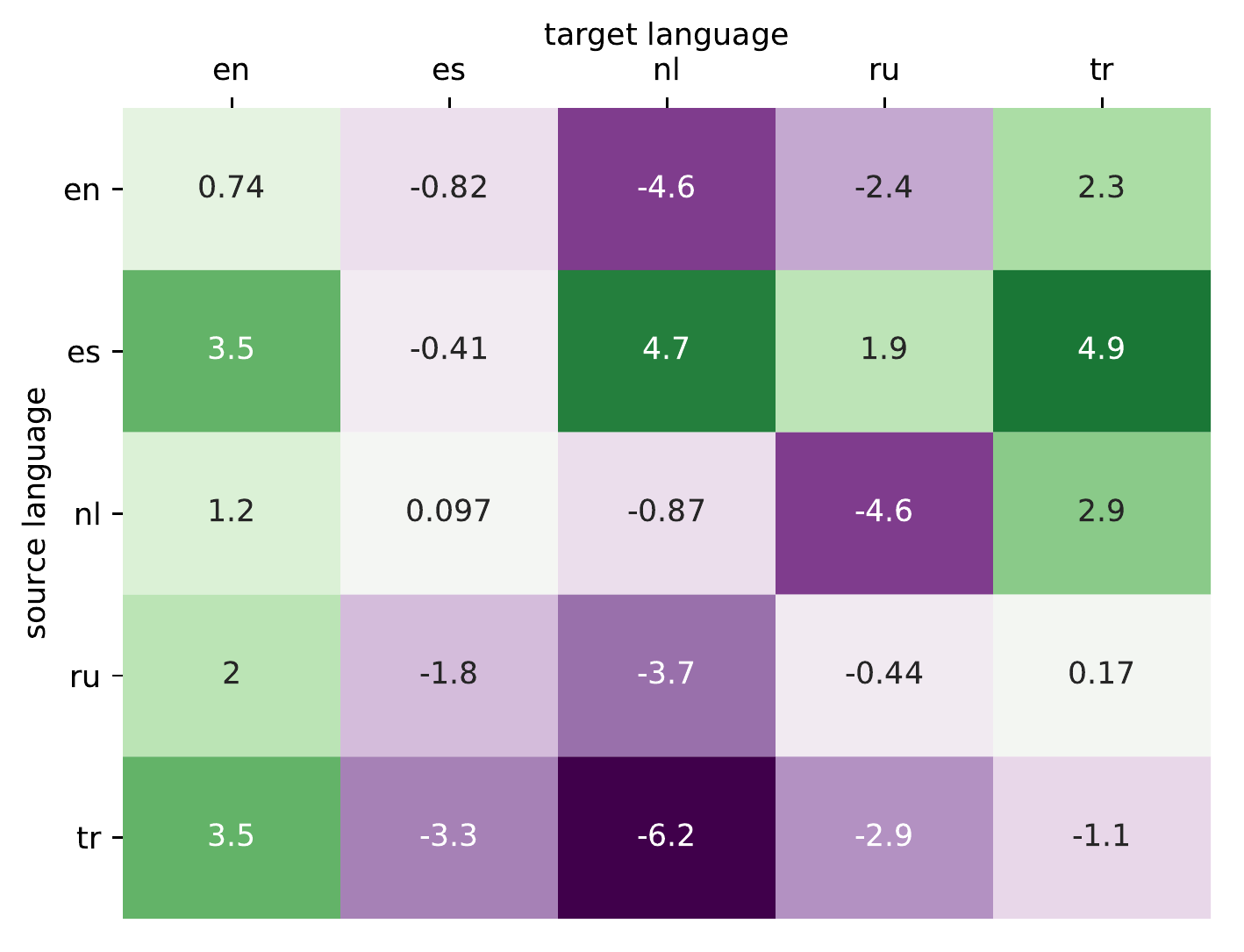}
  \caption{Zero-shot results comparing the multilingual embeddings \adv{} to \svd{}. A positive value means higher absolute F$_1$ score for \adv{} and vice versa. For readability, score differences are scaled by a factor of 100.}
  \label{fig:single-zero-diff}
\end{figure}

As can be seen in Figure~\ref{fig:single-zero-diff}, the two methods do perform well overall, albeit different for specific language pairs.
In a monolingual setting (i.e.~main diagonal), \adv{} performs slightly worse than \svd{} with the exception of \texttt{en$\rightarrow$en}.
Using \adv{}, Spanish appears to be a more effective source language than using \svd{} as the average performance is about 2.9 points higher.
It can also be observed that the cross-lingual transfer learning works better for English as a target language using \adv{} since the average performance is about 2.2 points higher than for \svd{}.
The opposite is true for Dutch as a target language, which shows a reduction in performance by 2.1 points on average.
Overall, for 13 of the 25 language pairs, the embeddings based on \svd{} perform better than embeddings aligned with \adv{}.

\subsection{Comparison to State-of-the-Art}
\label{sec:sota}
In this last part of our evaluation, we want to put our work into perspective of prior systems for opinion target extraction on the SemEval 2016 restaurant datasets.
We report results for our multilingual model that is trained on the combined training data of all languages and evaluated on the corresponding test datasets.
We compare our model to the respective state-of-the-art for each language in Table~\ref{tab:sota}.

\begin{table}[]
\small
\tabcolsep=0.11cm
\begin{tabular}{@{}llllll@{}}
\toprule
System                                          & \texttt{en} & \texttt{es} & \texttt{nl} & \texttt{ru} & \texttt{tr} \\ \midrule
\citet{Toh2016}                                 & 0.723       & --          & --          & --          & --          \\
\scriptsize{\citet{GTI}}                        & 0.666       & 0.685       & --          & --          & --          \\
\citet{kumar2016iit}                            & 0.685       & \textbf{0.697}       & \textbf{0.644}       & --          & --          \\
\citet{Pontiki2016}*                            & 0.441       & 0.520       & 0.506       & 0.493       & 0.419       \\
\citet{D17-1310}                                & \textbf{0.734}       & --          & --          & --          & --          \\
\texttt{all}$\rightarrow$\texttt{target} (Ours) & 0.660       & 0.687       & 0.624       & \textbf{0.567}       & \textbf{0.490}       \\ \bottomrule
\end{tabular}
\caption{Overview of the current state-of-the-art for opinion target extraction for 5 languages. Our model is trained on the combined training data of all languages and evaluated on the respective test datasets. The row marked with * is the baseline provided by the workshop organizers. To our knowledge, no better model is published for Russian and Turkish.}
\label{tab:sota}
\end{table}

We can see that the competition is strongest for English where we fall behind recent monolingual systems.
This corresponds to rank 7 of 19 of the original SemEval competition.
Regarding the other languages, we see that we are close to the best Spanish and Dutch systems and even clearly outperform systems for Russian and Turkish by at least 7 points in F$_1$-score.
With that, we present the first approach on this task to achieve such competitive performances for a variety of languages with a single, multilingual model.

\subsection{Discussion and Future Work}
The presented experiments shed light on the performance of our proposed approach under various circumstances.
In the following, we want to discuss its limitations and consider explanations for performance differences of different language pairs.

\paragraph{Model Limitations}
The core of our proposed sequence labeling approach consists of aligned word embeddings and shared CNN layers.
Due to the limited context of a CNN layer, the model can only base its decisions for each word on the local information around that word.
In many cases, this information is sufficient since most opinion target expressions are adjective-noun phrases\footnote{90\% of OTEs in the English dataset consist of zero or more adjectives followed by at least one noun.} which are well enough identified by the local context for most considered languages.

As future work, it is worth to investigate in how far our findings translate to more complex model architectures that have been proposed for OTE extraction, such as memory networks or attention-based models.

\paragraph{Language Characteristics}
Due to the inherent variability of natural languages and of the used datasets, it is difficult to identify the exact reasons for the observed performance differences between language pairs.
However, we suspect that language features such as word order, inflection, or agglutination affect the \emph{compatibility} of languages.
As an example, Turkish is considered a highly agglutinative language, that is, complex words are composed by attaching several suffixes to a word stem. 
This sets it apart from the other 4 languages.
This language feature might present a difficulty in our approach since the appending of suffixes is not optimally reflected in the tokenization process and the used word embeddings.
An approach that performs alignment of languages on subword units might alleviate this problem and lead to performance gains for language pairs with similar inflection rules.

Syntactic regularities such as word order might also play a role in our transfer learning approach.
It is reasonable to assume that the CNN layers of our approach pick up patterns in the word order of a source language that are indicative of an opinion target expression, e.g. \emph{"the [NOUN] is good"}.
When applying such a model to a target language with drastically different word order regularities, these patterns might not appear as such in the target language.

For the considered languages, we see following characteristics:
Where English and Spanish are generally considered to follow a Subject-Verb-Object (SVO) order, Dutch largely exhibits a combination of SOV and SVO cases. Turkish and Russian are overall flexible in their word order and allow a variety of syntactic structures.
In the case of Turkish, its morphological and syntactic features seem to explain some of the relatively low results.
However, with the small sample of languages and the many potential influencing factors at play, we are aware that it is not possible to draw any strong conclusions.
Further research has to be conducted in this direction to answer open questions.

\section{Related Work}
\label{sec:related}
Our work brings together the domains of opinion target extraction on the one side and cross lingual learning on the other side.
In this section, we give a brief overview of both domains and point out parallels to previous work.

\paragraph{Opinion Target Extraction}
\citet{agerri2015elixa} present a system that addresses opinion target extraction as a sequence labeling problem based on a perceptron algorithm with token, word shape and clustering-based features.

\citet{Toh2014} propose a Conditional Random Field (CRF) as a sequence labeling model that includes a variety of features such as Part-of-Speech (POS) tags and dependency tree features, word clusters and features derived from the WordNet taxonomy.
The model is later improved using neural network output probabilities \cite{Toh2016} and achieved the best results on the SemEval 2016 dataset for English restaurant reviews.

\citet{Jakob2010} follow a very similar approach that addresses opinion target extraction as a sequence labeling problem using CRFs.
Their approach includes features derived from words, Part-of-Speech tags and dependency paths, and performs well in a single and cross-domain setting.

\citet{kumar2016iit} present a CRF-based model that makes use of a variety of morphological and linguistic features and is one of the few systems that submitted results for more than one language for the SemEval 2016 ABSA challenge.
The strong reliance on high-level NLP features, such as dependency trees, named-entity information and WordNet features restricts its wide applicability to resource-poor languages.

Among neural network models \citet{Poria2016} and \citet{jebbara2016aspect} use deep convolutional neural network (CNN) with Part-of-Speech (POS) tag features.
\citet{Poria2016} also extend their base model using linguistic rules.

\citet{Wang2017} use coupled multi-layer attentions to extract opinion expressions and opinion targets jointly.
This approach, however, relies on additional annotations for opinion expressions alongside annotations for the opinion targets.

\citet{D17-1310} propose two LSTMs with memory interaction to detect aspect and opinion terms.
In order to generate opinion expression annotations for the SemEval dataset, a sentiment lexicon is used in combination with high precision dependency rules.

For a more comprehensive overview of ABSA and OTE extraction approaches we refer to \citet{Pontiki2016}.

\paragraph{Cross-Lingual and Zero-Shot Learning for Sequence Labelling}
With the CLOpinionMiner, \citet{zhou2015clo} present a method for cross-lingual opinion target extraction that relies on machine translation.
The approach derives an annotated dataset for a target language by translating the annotated source language data.
Part-of-Speech tags and dependency path-features are projected into the translated data using the word alignment information of the translation algorithm.
The approach is evaluated for English to Chinese reviews.
A drawback of the presented method is that it requires access to a strong machine translation algorithm for source to target language that also provides word alignment information.
Additionally, it builds upon NLP resources that are not available for many potential target languages.

Addressing the task of zero-shot spoken language understanding (SLU), \citet{upadhyay2018almost} follow a similar approach as our work.
They use the aligned embeddings from \citet{Smith2017offline} in combination with a bidirectional RNN and target zero-shot SLU for Hindi and Turkish.
\\

\noindent
Overall, our work differs from the related work by presenting a simple model for the zero-shot extraction of opinion target expressions.
By using no annotated target data or elaborate NLP resources, such as Part-of-Speech taggers or dependency parsers, our approach is easily applicable to many resource-poor languages.

\section{Conclusion}
\label{sec:conclusion}
In this work, we presented a method for cross-lingual and zero-shot extraction of opinion target expressions which we evaluated on 5 languages.
Our approach uses multilingual word embeddings that are aligned into a single vector space to allow for cross-lingual transfer of models.

Using English as a source language in a zero-shot setting, our approach was able to reach an F$_1$-score of 0.50 for Spanish and 0.46 for Dutch.
This corresponds to relative performances of 74\% and 77\% compared to a baseline system trained on target language data.
By using multiple source languages, we increased the zero-shot performance to F$_1$-scores of 0.58 and 0.53, respectively, which correspond to 85\% and 87\% in relative terms.
We investigated the benefit of augmenting the zero-shot approach with additional data points from the target language.
Here, we observed that we can save several hundreds of annotated data points by employing a cross-lingual approach.
Among the 5 considered languages, Turkish seemed to benefit the least from cross-lingual learning in all experiments. The reason for this might be that Turkish is the only agglutinative language in the dataset.
Further, we compared two approaches for aligning multilingual word embeddings in a single vector space and found their results to vary for individual language pairs but to be comparable overall.
Lastly, we compared our multilingual model with the state-of-the-art for all languages and saw that we achieve competitive performances for some languages and even present the best system for Russian and Turkish.

\section*{Acknowledgement}
This work was supported in part by the H2020 project Pr\^et-\`a-LLOD under Grant Agreement number 825182.

\bibliography{main}
\bibliographystyle{acl_natbib}

\end{document}